\documentclass{article} %
\usepackage[final]{template}

\usepackage{microtype}
\usepackage{hyperref}
\usepackage{url}
\usepackage{graphicx}
\usepackage{wrapfig}
\usepackage{amssymb}
\usepackage{amsthm}
\usepackage{amsmath}
\usepackage{booktabs}
\usepackage{makecell}
\usepackage{multirow}
\usepackage{caption}
\usepackage[capitalize,noabbrev]{cleveref}
\usepackage{subcaption}
\usepackage{enumitem}
\usepackage{adjustbox}
\usepackage{pgfplots}
\usepackage{titletoc}
\usepackage[most]{tcolorbox} %
\tcbuselibrary{listings,skins,breakable}
\usepackage{listings}
\usepackage[T1]{fontenc}        %
\usepackage{fontawesome5}    %
\usepackage{multirow}
\usepackage{xcolor,colortbl}
\usepackage{arydshln}
\usepackage{svg}
\usepackage[frozencache]{minted}

\usepackage{siunitx}
\usepackage{lineno}\lstdefinestyle{pythonstyle}{
  language=Python,
  basicstyle=\ttfamily\footnotesize,
  numbers=left,
  numberstyle=\ttfamily\scriptsize\color{gray!70},
  stepnumber=1,
  numbersep=10pt,          %
  xleftmargin=2.2em,       %
  framexleftmargin=2.2em,  %
  showstringspaces=false,
  keepspaces=true,
  breaklines=true,
  tabsize=4,
  columns=fullflexible,
  keywordstyle=\color{purple!80!black}\bfseries,
  commentstyle=\color{gray!60}\itshape,
  stringstyle=\color{green!50!black}
}

\lstdefinestyle{convstyle}{
    backgroundcolor=\color{lightgray},
    basicstyle=\ttfamily\scriptsize,
    breakatwhitespace=false,
    breaklines=true,
    captionpos=b,
    keepspaces=true,
    numbersep=5pt,
    showspaces=false,
    showstringspaces=false,
    showtabs=false,
    tabsize=2,
    showstringspaces=false,
    moredelim=**[is][\color{black}]{user}{user},
    moredelim=**[is][\color{blue}]{assistant}{assistant},
}

\lstdefinestyle{snippet}{
basicstyle=\ttfamily\footnotesize,
columns=fullflexible,
keepspaces=true,
breaklines=true,
breakatwhitespace=false,
tabsize=2,
showstringspaces=false,
showspaces=false,
showtabs=false,
captionpos=b,
float=false,
}

\usepackage{algorithm}
\usepackage{algpseudocode}

\algrenewcommand\algorithmiccomment[1]{\hfill{\footnotesize\(\triangleright\)~#1}}
\algrenewcommand\algorithmicrequire{\textbf{Input:}}
\algrenewcommand\algorithmicensure{\textbf{Output:}}

\definecolor{lightgray}{gray}{0.95}
\definecolor{green}{HTML}{59BB2B}

\usepackage{titlesec}
\titlespacing*{\paragraph}{0pt}{0.25ex}{2ex}

\hypersetup{
    colorlinks,
    linkcolor={blue!80!black},
    citecolor={blue!80!black},
    urlcolor={blue!80!black}
}
\setenumerate{left=12pt,itemsep=0pt,topsep=0pt}
\setitemize{left=12pt,itemsep=0pt,topsep=0pt}
\NewDocumentCommand{\incplt}{O{\columnwidth}m}{%
  \begin{center}
    \adjustbox{center}{\adjustbox{width=#1+10pt}{\includegraphics[width=#1]{./plots/output/#2.pdf}}}
  \end{center}
}
\newcommand{\figref}[2]{Figure~\hyperref[#1]{\ref{#1}~(#2)}}

\title{Anatomy of a Lie: A Multi-Stage Diagnostic Framework for Tracing Hallucinations in Vision-Language Models}

\author{%
Lexiang Xiong\textsuperscript{1}\thanks{Equal contribution.}, \, 
Qi Li\textsuperscript{1}\footnotemark[1], \,
Jingwen Ye\textsuperscript{2}, 
Xinchao Wang\textsuperscript{1}\thanks{Corresponding author.}\\
\textsuperscript{1}National University of Singapore\\
\textsuperscript{2}Monash University\\
\{lexiang, liqi\}@u.nus.edu, xinchao@nus.edu.sg\\[2ex]
\url{https://github.com/Lexiang-Xiong/CAD}
}

\crefname{appendix}{Appendix}{Appendices}
\Crefname{appendix}{Appendix}{Appendices}

\usepackage{amsmath}
\usepackage{amssymb}
\usepackage{mathtools}
\usepackage{amsthm}
\usepackage{bbm}
\usepackage{xspace}
\usepackage{nicefrac}
\usepackage{algorithm}
\usepackage{algpseudocode}  %
\usepackage{wrapfig}
\usepackage{subcaption}
\usepackage{titletoc}
\usepackage{multirow}
\usepackage{xcolor,colortbl}
\usepackage{arydshln}
\usepackage[framemethod=tikz]{mdframed}
\definecolor{lightgray}{gray}{0.95}

\usepackage[capitalize,noabbrev]{cleveref}

\newcounter{insight}

\usepackage{titlesec}
\titlespacing*{\paragraph}{0pt}{0.25ex}{2ex}

\definecolor{chaptercolor}{HTML}{1A254B}
\definecolor{darkblue}{HTML}{1A254B}
\definecolor{linkcolor}{HTML}{2B50AA}
\definecolor{citecolor}{HTML}{2B50AA}
\definecolor{lightlinkcolor}{HTML}{9A8F97}
\definecolor{darklinkcolor}{HTML}{1A254B}
\definecolor{light}{HTML}{F8F8F8}
\definecolor{lightblue}{HTML}{A7BED3}
\definecolor{red}{HTML}{F2545B}
\definecolor{blue}{HTML}{2b50aa}

\theoremstyle{plain}

\crefname{proposition}{Proposition}{Propositions}

\newtcolorbox[auto counter]{takeaway}[1][]{%
  enhanced,
  breakable,
  colback=black!5,          %
  colframe=black!85,        %
  boxrule=1.25pt,
  arc=3pt,                  %
  left=2mm,right=2mm,top=2mm,bottom=1.3mm,
  before skip=10pt, after skip=10pt,
  title={Takeaway~\thetcbcounter},
  colbacktitle=black!85,    %
  coltitle=white,           %
  fonttitle=\bfseries\small,
  attach boxed title to top left={yshift=-1.2mm, xshift=2mm},
  boxed title style={
    enhanced,
    arc=3pt,
    top=0.5mm, bottom=0.5mm, left=1mm, right=1mm,
    boxrule=0pt,           %
    interior engine=empty, %
  },
  #1                        %
}
\newtcolorbox{block}[1][]{%
  enhanced,
  breakable,
  colback=black!5,          %
  colframe=black!85,        %
  boxrule=1.25pt,
  arc=3pt,                  %
  left=2mm,right=2mm,top=1mm,bottom=1mm,
  before skip=10pt, after skip=10pt,
  #1                        %
}

\tcbset{
  chat_container/.style={
    enhanced, breakable,
    colback=gray!2!white,
    colframe=gray!50!black,
    arc=3mm,
    boxrule=0.6pt,
    left=2mm, right=2mm, top=2mm, bottom=2mm,
    fonttitle=\bfseries,
    title={Qualitative Example}
  }
}

\tcbset{
  problem_box/.style={
    enhanced, breakable,
    colback=teal!5!white,
    colbacklower=white,
    colframe=teal!50!black,
    arc=2mm,
    boxrule=0.6pt,
    left=2mm, right=2mm, top=1mm, bottom=1mm,
    fonttitle=\bfseries,
    title={Problem}
  },
  groundtruth_box/.style={
    enhanced, breakable,
    colback=green!5!white,
    colbacklower=white,
    colframe=green!50!black,
    arc=2mm,
    boxrule=0.6pt,
    left=2mm, right=2mm, top=1mm, bottom=1mm,
    fonttitle=\bfseries,
    title={Ground Truth}
  },
  reasoning_box/.style={
    enhanced, breakable,
    colback=blue!5!white,
    colbacklower=white,
    colframe=blue!50!black,
    arc=2mm,
    boxrule=0.6pt,
    left=2mm, right=2mm, top=1mm, bottom=1mm,
    fonttitle=\bfseries,
    title={Initial Answer}
  },
  finalanswer_box/.style={
    enhanced, breakable,
    colback=violet!5!white,
    colbacklower=white,
    colframe=violet!50!black,
    arc=2mm,
    boxrule=0.6pt,
    left=2mm, right=2mm, top=1mm, bottom=1mm,
    fonttitle=\bfseries,
    title={Final Answer}
  }
}

\NewDocumentCommand{\norm}{sm}{\IfBooleanTF{#1}{\|#2\|}{\left\| #2 \right\|}}
\NewDocumentCommand{\normF}{sm}{\IfBooleanTF{#1}{\|#2\|_{\mathrm{F}}}{\left\| #2 \right\|_{\mathrm{F}}}}
\NewDocumentCommand{\dTV}{sm}{d_{\mathrm{TV}}\IfBooleanTF{#1}{(#2)}{\left( #2 \right)}}

\DeclarePairedDelimiter\parentheses{(}{)}
\DeclarePairedDelimiter\brackets{[}{]}
\DeclarePairedDelimiter\braces{\{}{\}}

\NewDocumentCommand{\irred}{som}{\ensuremath{\sigma_{\hspace{-1pt}\infty}\IfBooleanTF{#1}{^2}{}(#3\IfValueTF{#2}{;#2}{})}}

\NewDocumentCommand{\fnPr}{}{\mathbb{P}}
\RenewDocumentCommand{\Pr}{om}{\fnPr\IfValueT{#1}{_{#1}}\parentheses*{#2}}
\RenewDocumentCommand{\H}{mo}{\mathrm{H}\IfValueTF{#2}{\!\left[#1\ \middle|\ #2\right]}{\brackets*{#1}}}
\NewDocumentCommand{\Hsm}{mo}{\mathrm{H}\IfValueTF{#2}{[#1 \mid #2]}{\brackets{#1}}}
\NewDocumentCommand{\I}{mmo}{\mathrm{I}\IfValueTF{#3}{\!\left(#1;#2\ \middle|\ #3\right)}{\parentheses*{#1; #2}}}
\NewDocumentCommand{\Ism}{mmo}{\mathrm{I}\IfValueTF{#3}{(#1;#2 \mid #3)}{\parentheses{#1; #2}}}

\NewDocumentCommand{\E}{somo}{\ensuremath{\mathbb{E}\IfValueT{#2}{_{#2}}{} \IfBooleanTF{#1}{#3}{\IfValueTF{#4}{\!\left[#3\ \middle|\ #4\right]}{\brackets*{#3}}}}}
\NewDocumentCommand{\Esm}{somo}{\ensuremath{\mathbb{E}\IfValueT{#2}{_{#2}}{} \IfBooleanTF{#1}{#3}{\IfValueTF{#4}{\!\left[#3\ \middle|\ #4\right]}{\brackets{#3}}}}}
\NewDocumentCommand{\Var}{somo}{\mathrm{Var}\IfValueT{#2}{_{#2}}{} \IfBooleanTF{#1}{#3}{\IfValueTF{#4}{\!\left(#3\ \middle|\ #4\right)}{\parentheses*{#3}}}}
\NewDocumentCommand{\Varsm}{somo}{\mathrm{Var}\IfValueT{#2}{_{#2}}{} \IfBooleanTF{#1}{#3}{\IfValueTF{#4}{\left(#3\ \middle|\ #4\right)}{\parentheses{#3}}}}
\NewDocumentCommand{\Cov}{som}{\mathrm{Cov}\IfValueT{#2}{_{#2}}{} \IfBooleanTF{#1}{#3}{\brackets*{#3}}}
\NewDocumentCommand{\Cor}{som}{\mathrm{Cor}\IfValueT{#2}{_{#2}}{} \IfBooleanTF{#1}{#3}{\brackets*{#3}}}

\NewDocumentCommand{\grad}{e_}{\boldsymbol{\nabla}\IfValueT{#1}{_{\!\!#1}\,}}

\NewDocumentCommand{\diag}{som}{\mathrm{diag}\IfValueT{#2}{_{#2}}{} \IfBooleanTF{#1}{\braces{#3}}{\braces*{#3}}}

\NewDocumentCommand{\N}{somm}{\mathcal{N}\IfBooleanTF{#1}{\left(}{(}\IfValueT{#2}{#2;}{} #3, #4\IfBooleanTF{#1}{\right)}{)}}
\NewDocumentCommand{\GP}{omm}{\mathcal{GP}(\IfValueT{#1}{#1;}{} #2, #3)}

\begin{document}

\maketitle

\vspace{-1ex}
\begin{abstract}

Vision-Language Models (VLMs) frequently `hallucinate'—generate plausible yet factually incorrect statements—posing a critical barrier to their trustworthy deployment. In this work, we propose a new paradigm for diagnosing hallucinations, recasting them from static output errors into dynamic pathologies of a model's computational cognition. Our framework is grounded in a normative principle of computational rationality, allowing us to model a VLM's generation as a dynamic \textit{cognitive trajectory}. We design a suite of information-theoretic probes that project this trajectory onto an interpretable, low-dimensional \textbf{Cognitive State Space}. Our central discovery is a governing principle we term the \textbf{geometric-information duality}: a cognitive trajectory's geometric abnormality within this space is fundamentally equivalent to its high information-theoretic surprisal. Hallucination detection is thus elegantly re-framed as a geometric anomaly detection problem. Evaluated across diverse settings—from rigorous binary QA (POPE) and comprehensive reasoning (MME) to unconstrained open-ended captioning (MS-COCO)—our framework achieves state-of-the-art performance. Crucially, it operates with high efficiency under weak supervision and remains highly robust even when calibration data is heavily contaminated. This approach enables a causal attribution of failures, mapping observable errors to distinct pathological states: \textbf{perceptual instability} (measured by Perceptual Entropy, $H_{\text{Evi}}$), \textbf{logical-causal failure} (measured by Inferential Conflict, $S_{\text{Conf}}$), and \textbf{decisional ambiguity} (measured by Decision Entropy, $H_{\text{Ans}}$). Ultimately, this opens a path toward building AI systems whose reasoning is transparent, auditable, and diagnosable by design.
  \looseness=-1
\end{abstract}

\startcontents

\section{Introduction}
\label{sec:introduction}
\begin{figure*}[t]
    \centering
    
    \includegraphics[width=1.0\linewidth]{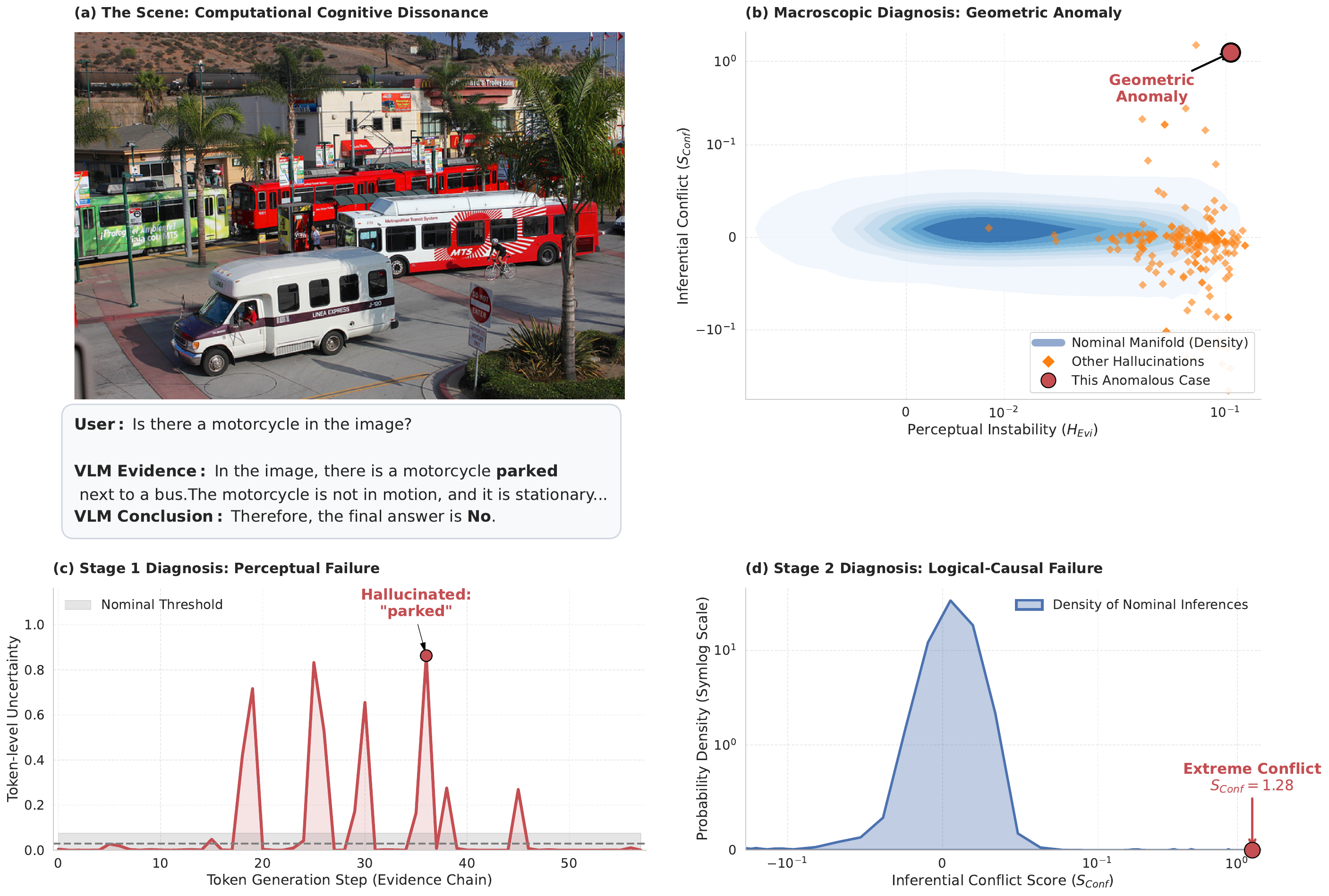} 
    \caption{
    An example of \textit{computational cognitive dissonance} in Idefics2, where a cascade of failures leads to a coincidentally correct answer.
    \textbf{(1) Perceptual Failure:} The model hallucinates a `motorcycle' in the evidence chain, an object not present in the image (a cyclist is visible). Our framework captures this as high \textbf{Perceptual Instability} (see panel (c)).
    \textbf{(2) Logical Failure:} The model then contradicts its own faulty evidence, concluding the final answer is `No'. This breakdown of self-consistency is diagnosed as extremely high \textbf{Inferential Conflict} (see panel (d)).
    This case study demonstrates the limitation of accuracy-only evaluations and highlights our framework's ability to perform a stage-by-stage differential diagnosis of a VLM's cognitive process, identifying complex, multi-stage failure trajectories.
}
\vspace{-7mm}
\label{fig:figure1}
\end{figure*}

Consider a striking paradox in Vision-Language Models (VLMs)~\cite{yu2025discrete,li2025vid,qwen2vl,li2025every,Li2023Evaluating}: when asked ``Is there a motorcycle in the image?'', a model might confidently hallucinate evidence—``In the image, there is a motorcycle parked''—yet inexplicably conclude, ``Therefore, the final answer is No.'' We term this cascade of errors \textit{computational cognitive dissonance}, as illustrated in \Cref{fig:figure1}. What makes this specific case so dramatic is that a double failure (perceiving a non-existent object, then logically contradicting that very perception) leads to a \textit{coincidentally correct} final answer.

This phenomenon exposes a critical insight that severely limits VLM deployment in high-stakes domains~\cite{Ji2023Survey,li2026sponge,Bai2024Hallucination,li2024data,wang2025towards}: hallucinations are rarely monolithic errors that can be diagnosed by a single metric like ``accuracy'' or ``self-consistency.'' Instead, they are often complex, \textbf{multi-stage pathologies} where distinct failures—such as perceptual drift and logical bypass—compound and interact within a single cognitive trajectory.
Current approaches to hallucination detection generally treat the generation process as an indivisible, monolithic event. They either evaluate the semantic consistency of final outputs via multiple sampling \cite{Manakul2023SelfCheckGPT, Farquhar2024Detecting} or probe for a binary `truthfulness' representation within internal states \cite{Azaria2023The, Chen2024INSIDE}. While foundational, these reductionist views conflate fundamentally different failure modes. They struggle to distinguish whether a hallucination stems from an initial failure to ground concepts in the image (\textit{perceptual drift}) or from an illogical jump that bypasses extracted facts (\textit{inferential bypass}). Our central thesis is that hallucination is a process-level failure that must be diagnosed within a structured model of cognition.

To address this, we introduce a normative principle of computational rationality~\cite{gershman2015computational,oulasvirta2022computational} for VLMs, formalized as a Markovian information flow: Image ($\mathcal{I}$) $\xrightarrow{\text{Perception}}$ Textual Evidence ($\mathcal{T}_{\text{evi}}$) $\xrightarrow{\text{Inference}}$ Final Answer ($\mathcal{A}$). This principle asserts that for a rational agent, the final answer $\mathcal{A}$ is conditionally independent of the image $\mathcal{I}$ given the evidence $\mathcal{T}_{\text{evi}}$, implying the conditional mutual information $I(\mathcal{A}; \mathcal{I} | \mathcal{T}_{\text{evi}})$ must be zero. Critics might argue that requiring an explicit evidence chain limits the applicability of such a framework. However, we employ Chain-of-Thought (CoT) not as a strict operational constraint, but as a crucial \textbf{diagnostic probe in explainable AI (XAI)}—akin to a medical contrast agent. By forcing the model to externalize its latent reasoning, we make the implicit cognitive trajectory observable and mathematically diagnosable.

To diagnose this cognitive process, we design a suite of probes. While $S_{\text{Conf}}$ directly measures violations of our core principle, Perceptual Entropy ($H_{\text{Evi}}$) and Decision Entropy ($H_{\text{Ans}}$) quantify the stability of the process's initial and final stages, providing a complete diagnostic picture. These probes act as natural coordinates to project the high-dimensional trajectory onto an interpretable 3D \textbf{Cognitive State Space}:
\begin{itemize}[leftmargin=*, topsep=2pt, itemsep=-2pt]
    \item \textbf{Perceptual Instability ($H_{\text{Evi}}$):} Measured via Perceptual Entropy, this probes the uncertainty at the perception stage ($\mathcal{I} \rightarrow \mathcal{T}_{\text{evi}}$).
    \item \textbf{Logical-Causal Failure ($S_{\text{Conf}}$):} Measured via Inferential Conflict, this directly quantifies the information leakage that violates our core principle.
    \item \textbf{Decisional Ambiguity ($H_{\text{Ans}}$):} Measured via Decision Entropy, this probes the final uncertainty at the trajectory's terminal stage.
\end{itemize}
Collectively, these probes summarize each cognitive trajectory as a Cognitive State Vector within this space. 

This perspective reveals a powerful \textbf{geometric-information duality}: a trajectory's geometric abnormality within this space is fundamentally an expression of its high information-theoretic surprisal. Our experimental results (see \cref{fig:density_manifolds}) provide strong empirical evidence for this duality. Normative cognitive trajectories consistently evolve towards stable, low-energy basins of attraction, forming a dense submanifold. Hallucinations, conversely, are high-energy deviations that are perturbed off this manifold, appearing as geometric anomalies. This duality serves as the theoretical bridge that translates the semantic problem of hallucination into a rigorous geometric anomaly detection task~\cite{stolz2020geometric}.

This novel reframing achieves state-of-the-art detection performance while offering significant practical advantages. Unlike multi-sample methods, our approach requires only a single generation pass plus a highly efficient non-autoregressive replay (detailed in \cref{sec:method}). Furthermore, it operates under weak supervision (requiring only ground-truth answers, not fine-grained hallucination labels) and remains highly resilient even when calibration data is contaminated with up to 30\% noise. To demonstrate its versatility, we rigorously validate our framework across a spectrum of tasks: from the controlled adversarial subset of POPE \cite{Li2023Evaluating}, to the comprehensive reasoning categories of MME \cite{fumme}, and finally to unconstrained open-ended captioning on MS-COCO \cite{lin2014microsoft}. 

In summary, our primary contribution is not merely a new state-of-the-art hallucination detector, but a principled diagnostic framework that provides a new lens through which to understand VLM failures. Our contributions are:
\begin{itemize}[leftmargin=*, topsep=2pt, itemsep=-2pt]
    \item We propose a diagnostic framework that reframes hallucination from a static flaw to a dynamic analysis of a VLM's cognitive trajectory, grounded in a normative principle of computational rationality.
    \item We design a suite of information-theoretic probes that act as natural coordinates to project the generative process onto an interpretable \textbf{Cognitive State Space}, enabling a stage-by-stage differential diagnosis.
    \item Grounded in a powerful geometric-information duality, we introduce a novel detection method based on geometric anomaly detection, which operates under weak supervision and achieves state-of-the-art performance.
    \item We deliver a novel mechanistic categorization of VLM failure modes by analyzing the topology of their cognitive manifolds, diagnosing complex errors like `computational cognitive dissonance'.
\end{itemize}
\section{Related Work}
\label{sec:related_work}

Our research is positioned at the confluence of VLM hallucination evaluation, mitigation, and internal state analysis \cite{Bai2024Hallucination, Ji2023Survey}. With these foundations, we introduce a novel \textit{diagnostic} framework that moves beyond static error detection to model the VLM's dynamic internal cognitive process.

\paragraph{VLM Hallucination Benchmarking.}

A significant body of work has focused on quantifying VLM hallucinations from final outputs. The pioneering CHAIR metric \cite{rohrbach2018object} measured hallucinated objects in captions. To address its instability, works like POPE \cite{Li2023Evaluating} and ROPE \cite{Chen2024MultiObject} established a stable polling-based evaluation paradigm. To capture more complex failure modes, recent benchmarks have expanded to evaluate real-world instruction following (VisIT-Bench \cite{bitton2023visit}, MMHal-Bench \cite{sun2023aligning}, HallusionBench \cite{guan2023hallusionbench}), fine-grained visual-text alignment (LM2-Bench \cite{peyrard2024lm2}, WYSWIR \cite{wang2023what}), and advanced commonsense reasoning (Visual Riddles \cite{yarom2024visual}). Adversarial benchmarks such as MAD-Bench \cite{Qian2024How} further probe robustness against deceptive prompts. While these static evaluation suites are invaluable for assessing \textit{what} errors a model makes across diverse scenarios, they predominantly treat the VLM as a black box. Our work provides a crucial complement: diagnosing the dynamic generative process itself to explain \textit{how} and \textit{why} these errors occur.

\paragraph{Inference-Time Hallucination Detection and Mitigation.}
Recent efforts have focused on inference-time strategies. One prominent line of work is contrastive decoding, which penalizes outputs driven primarily by language priors rather than visual evidence \cite{Wang2025TPC, Vu2025HalluField}. State-of-the-art methods like Hallucination-Induced Optimization (HIO) \cite{Chen2024Alleviating} refine this by training a dedicated `evil' model to provide a targeted contrastive signal. Parallel efforts in automated evaluation (auto-eval) employ strong LLMs (e.g., Clair \cite{tsun2024clair}) or contrastive grounding techniques (e.g., Contrastive Region Guidance \cite{wang2024contrastive}) to assess output quality without manual labels. Another direction analyzes internal signals, such as VADE \cite{Prabhakaran2025VADEVA}, which models attention map sequences. While these methods excel at scoring or \textit{correcting} the final output, our framework is distinctly focused on \textit{diagnosing the mechanistic failure}. Our Inferential Conflict metric (\cref{sec:method}) directly isolates the illicit vision-language information flow, offering a causal interpretation that auto-evals typically lack.

\paragraph{Internal State Analysis for Hallucination.}
A nascent line of inquiry explores VLM internal states, inspired by seminal research in LLMs suggesting truthfulness is encoded in hidden activations \cite{Azaria2023The, Chen2024INSIDE, Orgad2024LLMs, Ferrando2024Do}. While methods like VADE \cite{Prabhakaran2025VADEVA} analyze internal patterns, they focus on attention mechanisms. Other approaches, often adapted from the text-only domain, may treat the VLM's internal state as a monolithic representation \cite{Du2024HaloScope, Park2025Steer}. This simplification is ill-equipped to distinguish between a failure in initial perception versus a breakdown in subsequent reasoning. \textbf{Distinct from all prior work}, our research introduces a multi-faceted diagnostic framework that models a VLM's reasoning not as a static state, but as a \textbf{measurable cognitive trajectory through distinct, macroscopic stages}. This process-oriented view enables a mechanistic, differential diagnosis of where a breakdown originates.

\paragraph{Practical Advantages of Our Framework.}
Beyond its theoretical grounding, our framework offers significant practical advantages. It operates under weak supervision—requiring only ground-truth final answers rather than expensive, often ambiguous token-level annotations required by fully supervised detectors. Once calibrated, our method is highly efficient, requiring only the initial generation and a single non-autoregressive forward pass through the language decoder. This makes it significantly faster than multi-sample consistency methods \cite{Manakul2023SelfCheckGPT, Farquhar2024Detecting} and highly scalable for real-world deployment.
\section{Methodology: An Information-Geometric Framework for Diagnosing Hallucination}
\label{sec:method}

We reconceptualize VLM hallucination not as a simple output error, but as a symptom of a breakdown within the model's internal information processing. We first formalize the ideal, logically self-consistent cognitive process through an axiomatic probabilistic graphical model (PGM) that defines the normative flow of information: $\mathcal{I} \xrightarrow{\text{Perception}} \mathcal{T}_{\text{evi}} \xrightarrow{\text{Inference}} \mathcal{A}$, where $\mathcal{I}$ is the visual input, $\mathcal{T}_{\text{evi}}$ is the explicitly generated evidence chain, and $\mathcal{A}$ is the final answer to a given query $\mathcal{Q}$. 

This model embodies a critical axiom from information theory: the generated evidence $\mathcal{T}_{\text{evi}}$ serves as a \textbf{sufficient statistic} for the final answer $\mathcal{A}$ with respect to the image $\mathcal{I}$. Mathematically, this defines the ground truth of a rational process as one where the conditional mutual information is zero: $I(\mathcal{A}; \mathcal{I} | \mathcal{T}_{\text{evi}}) = 0$. Deviations from this axiom signify a logical failure. While our conceptual framework applies to general generation, we anchor our mathematical formalization and primary diagnosis in structured Visual Question Answering (VQA) tasks, as their constrained action spaces allow for rigorous quantification of these latent failure modes. 

\paragraph{An Information-Geometric View of Cognition.}
We define a VLM's generation of a token trajectory $\tau$ as a probabilistic event drawn from a distribution $P(\tau | \mathcal{I}, Q)$. The informational content of any specific trajectory is its \textbf{self-information}, or \textbf{surprisal}, defined as $I(\tau) = -\log P(\tau | \mathcal{I}, Q)$. This allows for a rigorous, first-principles definition of hallucination: a \textit{nominal} cognitive process is a low-surprisal event, corresponding to a high-probability trajectory that aligns with the model's learned world model. Conversely, we define hallucination as a \textbf{high-surprisal cognitive event}—a rare, low-probability trajectory that deviates unexpectedly from this nominal behavior.

To diagnose such events, we project the high-dimensional internal state of the generation process into a low-dimensional, 3D Observable Information Manifold. Each generation is represented by a Cognitive State Vector $v = [H_{\text{Evi}}, S_{\text{Conf}}, H_{\text{Ans}}]$ on this manifold. We posit that nominal processes correspond to points residing in high-density regions, or `attractors' on this manifold. Our three diagnostic probes are thus reinterpreted as direct, quantitative measures of the information flow's properties at different stages:
\begin{itemize}[leftmargin=*]
    \item \textbf{Perceptual Entropy ($H_{\text{Evi}}$)} measures the \textit{initial state's information entropy}, quantifying the uncertainty in the evidence formulation stage.
    \item \textbf{Inferential Conflict ($S_{\text{Conf}}$)} directly measures \textit{information leakage} across cognitive stages, quantifying the violation of our core axiom.
    \item \textbf{Decision Entropy ($H_{\text{Ans}}$)} quantifies the \textit{terminal state's residual entropy}, measuring the final decision uncertainty.
\end{itemize}

\subsection{Probing Perceptual Uncertainty ($H_{\text{Evi}}$)}
\label{sec:metric1}
To quantify the \textit{initial state's information entropy}, we measure the uncertainty of the evidence formulation stage ($\mathcal{I} \rightarrow \mathcal{T}_{\text{evi}}$) with Perceptual Entropy ($H_{\text{Evi}}$).\footnote{The complete word lists, adapted from prior work on language model uncertainty \cite{Ji2025Calibrating, yona2024can}, along with a sensitivity analysis, are in Appendix~\ref{app:metric_calc}.} A high $H_{\text{Evi}}$ signifies an unstable starting point for the cognitive trajectory. We model the model's choice at each token step $i$ as a Bernoulli trial $C_i \in \{\text{Factual}, \text{Uncertain}\}$ by defining two disjoint token subsets: a factual set $V_f$ and an uncertainty set $V_u$. For the logits $\mathbf{l}_i$ of each token, we project the softmax distribution onto this semantic axis:
\begin{equation}
    p(C_i=\text{F}) = \frac{\sum_{t \in V_f} \text{softmax}(\mathbf{l}_i)_t}{\sum_{t \in V_f \cup V_u} \text{softmax}(\mathbf{l}_i)_t}, \quad p(C_i=\text{U}) = 1 - p(C_i=\text{F})
\end{equation}
The token-level entropy is the Shannon entropy $H_i = H(C_i)$. The final metric is the path-averaged entropy: $H_{\text{Evi}} = \frac{1}{|\mathcal{T}_{\text{evi}}|} \sum_{i=1}^{|\mathcal{T}_{\text{evi}}|} H_i$.

\subsection{Probing Inferential Conflict ($S_{\text{Conf}}$)}
\label{sec:metric2}
To operationalize our idealized causal graph ($\mathcal{I} \rightarrow \mathcal{T}_{\text{evi}} \rightarrow \mathcal{A}$), we introduce \textbf{Inferential Conflict ($S_{\text{Conf}}$)}. This probe estimates the \textit{Conditional Pointwise Mutual Information} (CPMI) to quantify the strength of the illicit direct causal path from $\mathcal{I}$ to $\mathcal{A}$. It is a pointwise metric for a specific outcome $\mathcal{A}_{\text{token}}$, making it highly suitable for diagnosing single, concrete instances of generation. It measures the information gain from the visual modality on the generated answer token $\mathcal{A}_{\text{token}}$\footnote{For multi-token answers, $\mathcal{A}_{\text{token}}$ is defined as the first token corresponding to the primary decision keyword (e.g., 'Yes' or 'No'). This localization is made reliable by the structured prompts used in our experimental setup.}, conditioned on the textual evidence $\mathcal{T}_{\text{evi}}$. This quantity is computed as the \textbf{log-probability difference}:
\begin{align}
S_{\text{Conf}}
&=\log p_{v}(\mathcal{A}_{\text{token}}|\mathcal{I},\mathcal{T}_{\text{evi}})
  -\log p_{t}(\mathcal{A}_{\text{token}}|\emptyset_{\mathcal{I}},\mathcal{T}_{\text{evi}}) \\
&=\text{CPMI}(\mathcal{A}_{\text{token}};\mathcal{I}\mid\mathcal{T}_{\text{evi}})
\end{align}
where $p_v$ is the probability with visual context and $p_t$ is the counterfactual probability without it. A large positive $S_{\text{Conf}}$ indicates strong, positive point-wise information flowing directly from the visual input to the final answer, unmediated by the evidence, thus measuring the violation of d-separation. To obtain $p_t$, we perform a causal intervention by replaying the generation process with the visual input ablated\footnote{In our implementation with Idefics2, this is achieved by providing `images=None' to the processor during the text-only forward pass.} \cite{Pearl2009Causality}. A practical boundary condition is that the VLM architecture must allow for such a causal intervention.

\begin{algorithm}[t]
\caption{Cognitive Anomaly Detection Framework (CAD)}
\label{alg:framework}
\label{alg:framework_highly_compact}
\begin{algorithmic}[1]
\Statex \textbf{Function} \textsc{DiagnoseHallucination} $(I, Q, \mathcal{M}, \text{CalibratedComponents})$
\State $(\mathcal{M}_{\text{GMM}}, \mu, \sigma) \leftarrow \text{CalibratedComponents}$
\State
\State \Comment{1. Generate cognitive trajectory and extract metrics}
\State $(\mathcal{T}_{\text{evi}}, \mathcal{A}_{\text{model}}, \text{scores}) \leftarrow \mathcal{M}.\text{generate}(I, Q, \text{output\_scores=True})$
\State $H_{\text{Evi}} \leftarrow \text{CalcPerceptualEntropy}(\text{scores}_{\text{evi}})$
\State $S_{\text{Conf}} \leftarrow \text{CalcInferentialConflict}(I, Q, \mathcal{T}_{\text{evi}}, \mathcal{M})$
\State $H_{\text{Ans}} \leftarrow \text{CalcDecisionEntropy}(\text{scores}_{\text{ans}})$
\State
\State \Comment{2. Compute anomaly score in the cognitive space}
\State $v_{\text{new}} \leftarrow [H_{\text{Evi}}, S_{\text{Conf}}, H_{\text{Ans}}]$
\State Standardize $v_{\text{new}}$ using $\mu, \sigma$.
\State $S_{\text{hall}} \leftarrow - \log p(v_{\text{new}} | \mathcal{M}_{\text{GMM}})$
\State \textbf{return} $S_{\text{hall}}$
\Statex
\Statex \textit{// Note: CalibratedComponents are pre-computed offline by fitting a GMM on
% }\Statex \textit{// 
cognitive state vectors from a purified set of non-hallucinatory samples.}
\end{algorithmic}
\end{algorithm}
\subsection{Probing Decision Uncertainty ($H_{\text{Ans}}$)}
\label{sec:metric3}
Finally, to quantify the \textit{terminal state's residual entropy}, we measure the final uncertainty with Decision Entropy ($H_{\text{Ans}}$). A high entropy indicates the system has failed to converge to a stable, determined state.
\begin{equation}
    H_{\text{Ans}} = - \sum_{a \in \{\text{Yes}, \text{No}\}} p(a)\log_2 p(a)
\end{equation}
\subsection{Diagnosis via Geometric Anomaly Detection in the Cognitive State Space}
\label{sec:gmm_diagnosis}
Our framework performs diagnosis at inference-time on single instances after a one-time, hallucination-label-free calibration. This phase learns the geometric structure of the `nominal cognitive state space' $\mathcal{S}_{\text{nominal}}$.

\paragraph{Phase 1: Learning the Geometry of the Nominal Cognitive State Space.} We represent each VLM generation by its 3D Cognitive State Vector $v = [H_{\text{Evi}}, S_{\text{Conf}}, H_{\text{Ans}}]$. This calibration is fitted on a calibration set $\mathcal{D}_{\text{cal}}$. This process requires only ground-truth final answers (e.g., `Yes'/`No'), a form of \textbf{weak supervision} that is vastly more accessible and scalable than obtaining fine-grained, token-level hallucination labels. We hypothesize that the landscape of nominal states is multi-modal, as different types of valid cognitive processes (e.g., simple object recognition versus complex relational reasoning) may form distinct, dense clusters in the state space. We therefore employ a Gaussian Mixture Model (GMM), which is naturally suited to capturing such underlying structures, to model the probability density $p(v | \text{nominal})$. This set undergoes a Coherence Filter step: we first select for correct final answers and then apply automated heuristics to exclude `lucky guesses' (e.g., cases where the model answers `Yes' while its generated evidence explicitly states `There is no such object in the image'). This purification ensures our GMM learns a less biased estimate of the true density on $\mathcal{S}_{\text{nominal}}$. Prior to fitting, we standardize each dimension and determine the optimal number of GMM components via the Bayesian Information Criterion (BIC).

\paragraph{Phase 2: Hallucination Diagnosis as a High-Surprisal Cognitive Event.} From an information geometry perspective, a hallucination is a cognitive process whose state vector $v$ is geometrically distant from the learned high-density regions (attractors). Its Hallucination Score is therefore the self-information content, or surprisal, of observing this atypical state vector:
\begin{equation}
    S_{\text{hall}}(v) = I(v) = -\log p(v | \mathcal{M}_{\text{GMM}})
\end{equation}
This score quantifies the `unexpectedness' of the observed cognitive trajectory. Nominal processes are common, predictable, low-information events, whereas hallucinations are rare, high-information deviations. The workflow is summarized in Algorithm~\ref{alg:framework}.

\section{Experiments}
\label{sec:experiments}

\subsection{Experimental Setup}
\label{sec:setup}

\begin{table}[t]
    \centering
    \caption{\textbf{Main AUC results on the POPE benchmark (Adversarial).} Best performance is in \textbf{bold}, second best is \underline{underlined}. Our single-pass, weakly supervised CAD significantly outperforms all baselines.}
    \label{tab:main_results}
    \resizebox{\columnwidth}{!}{%
    \begin{tabular}{l|c|cccc|c}
        \toprule
        \textbf{Method} & \textbf{Cost} & \textbf{Llava-v1.6} & \textbf{Idefics2} & \textbf{Qwen2-VL} & \textbf{DeepSeek-VL} & \textbf{Average} \\
        \midrule
        Token Entropy & 1x & 0.603 & 0.806 & 0.409 & \underline{0.732} & 0.638 \\
        Neg Log Prob & 1x & 0.604 & 0.832 & 0.428 & 0.710 & 0.644 \\
        Supervised Probe & 1x & \underline{0.787} & \underline{0.898} & \underline{0.762} & 0.715 & \underline{0.791} \\
        Semantic Entropy & 10x & 0.711 & 0.751 & 0.673 & 0.702 & 0.709 \\
        \textbf{Ours (CAD)} & \textbf{1x} & \textbf{0.910} & \textbf{0.947} & \textbf{0.776} & \textbf{0.798} & \textbf{0.858} \\
        \bottomrule
    \end{tabular}%
    }
\end{table}

\begin{table}[t]
    \centering
    \caption{\textbf{Macro-averaged AUC results on the MME benchmark.} Our CAD framework shows strong generalization across diverse multimodal reasoning and perceptual tasks.}
    \label{tab:mme_results}
    \resizebox{.9\columnwidth}{!}{%
    \begin{tabular}{l|ccccc}
        \toprule
        \textbf{Method} & \textbf{Llava-v1.6} & \textbf{Idefics2} & \textbf{Qwen2-VL} & \textbf{DeepSeek-VL} & \textbf{Average} \\
        \midrule
        Token Entropy & 0.5106 & 0.5486 & 0.5291 & 0.6687 & 0.5643 \\
        Neg Log Prob & 0.5061 & 0.5449 & 0.5136 & 0.6578 & 0.5556 \\
        Supervised Probe & \underline{0.7620} & \underline{0.7905} & \textbf{0.7408} & \underline{0.7104} & \underline{0.7509} \\
        Semantic Entropy & 0.6811 & 0.7081 & 0.6616 & 0.6199 & 0.6677 \\
        \textbf{Ours (CAD)} & \textbf{0.8514} & \textbf{0.8411} & \underline{0.7233} & \textbf{0.7680} & \textbf{0.7960} \\
        \bottomrule
    \end{tabular}%
    }
\end{table}
\paragraph{Datasets and Multi-Dimensional Evaluation Protocol.}
Our work's core philosophy is that object-level hallucination is merely the final, observable symptom of a broader cognitive failure. To thoroughly evaluate our framework across diverse settings and address the limitations of narrow benchmark testing, we design a multi-dimensional evaluation protocol:
\begin{itemize}[leftmargin=*, topsep=0pt, itemsep=0pt]
    \item \textbf{Diagnostic Deep-Dive (POPE~\cite{Li2023Evaluating}):} We transform the POPE benchmark into a rich diagnostic playground using Chain-of-Thought (CoT) prompts to externalize reasoning ($\mathcal{T}_{\text{evi}}$). Following~\cite{Li2023Evaluating}, we strictly focus on the `adversarial` subset to diagnose genuine, hard-to-detect hallucinations, which serves as our primary testbed for mechanistic analysis.
    \item \textbf{Comprehensive Generalization (MME~\cite{fumme}):} To ensure our method generalizes beyond specific task formats, we evaluate on the expansive MME benchmark, reporting macro-averaged results across its diverse perceptual and reasoning categories.
    \item \textbf{Open-Ended Validation (MS-COCO~\cite{lin2014microsoft}):} As a targeted ablation, we validate our perceptual probe on open-ended image captioning, using the CHAIR~\cite{rohrbach2018object} metric to confirm its independent generalizability.
\end{itemize}

\paragraph{Evaluated Models and Baselines.}
To ensure a comprehensive analysis across architectural families, we evaluate four state-of-the-art VLMs: Llava-v1.6-Mistral-7B~\cite{llava_v1_6_mistral_7b}, Idefics2-8b~\cite{laurencon2023obelics}, Qwen2-VL~\cite{qwen2vl}, and DeepSeek-VL2-Small~\cite{deepseek_vl}. We compare our proposed \textbf{Cognitive Anomaly Detection (CAD)} against a comprehensive suite of baselines: 1) \textbf{Token Entropy} and \textbf{Neg Log Probability} (token-level uncertainty); 2) a \textbf{Supervised Probe}~\cite{Chen2024INSIDE} (a linear classifier trained on final hidden states using balanced hallucination labels); and 3) \textbf{Semantic Entropy}~\cite{Farquhar2024Detecting} (a strong, sampling-based consistency baseline requiring $10\times$ inference cost). Detailed metric calculations, GMM hyperparameter optimization (via BIC), and prompt templates are provided in the Appendix.

\subsection{State-of-the-Art Detection Across Diverse Benchmarks}

We find that reframing hallucination as a diagnosable \textit{cognitive process anomaly} leads to a state-of-the-art framework that is both effective and efficient. 

\begin{wrapfigure}{r}{0.55\columnwidth}
    \centering
    \vspace{-4mm}
    \includegraphics[width=\linewidth]{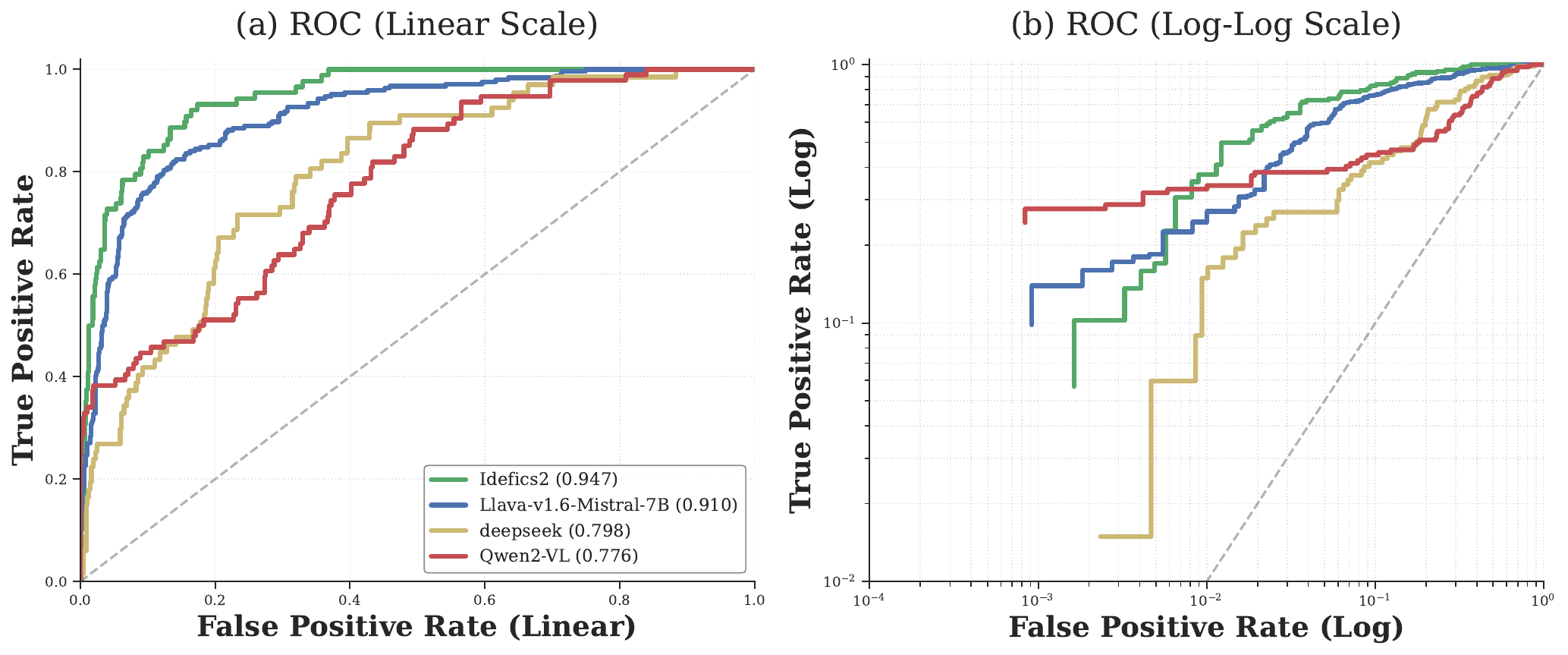}
    \caption{\textbf{ROC curves of our Cognitive Anomaly Detection (CAD) framework.} 
        (a) Linear-scale ROC curves show superior overall performance across all architectures. 
        (b) log-log curves highlight CAD's dominance in the critical low-FPR regime ($\text{FPR} < 10^{-2}$), which is essential for reliable real-world deployment.}
    \label{fig:roc_curves}
    \vspace{-4mm}
\end{wrapfigure}

\paragraph{Performance on POPE.} As shown in \cref{tab:main_results} and \cref{fig:roc_curves}, our CAD framework—which models the distribution of only non-hallucinatory examples— achieves superior overall performance. 
Critically, the log-log ROC curves (\cref{fig:roc_curves}b) highlight that CAD maintains high true positive rates even at extremely low false positive rates ($\text{FPR} < 10^{-2}$), a regime where baseline methods often fail. This confirms that modeling the geometric properties of the cognitive process offers superior reliability for real-world deployment compared to simple uncertainty thresholds.

\paragraph{Generalization on MME.} Moving beyond simple object existence questions, \cref{tab:mme_results} demonstrates CAD's strong generalization capabilities on the comprehensive MME benchmark. MME encompasses diverse multimodal tasks including spatial reasoning, OCR, and commonsense logic. In these complex scenarios, the natural variance in generated text increases significantly. While the Supervised Probe, benefiting from in-domain training labels, shows competitive performance, our weakly-supervised CAD framework achieves comparable or even superior results (e.g., 0.851 on Llava-v1.6) without requiring any hallucination labels for calibration. This highlights CAD's robustness and adaptability across a wide array of diverse task types.

\subsection{Mechanistic Diagnosis: Unveiling Cognitive Fingerprints}
\label{sec:diagnosis}

\begin{figure*}[t]
    \centering
    \includegraphics[width=\textwidth]{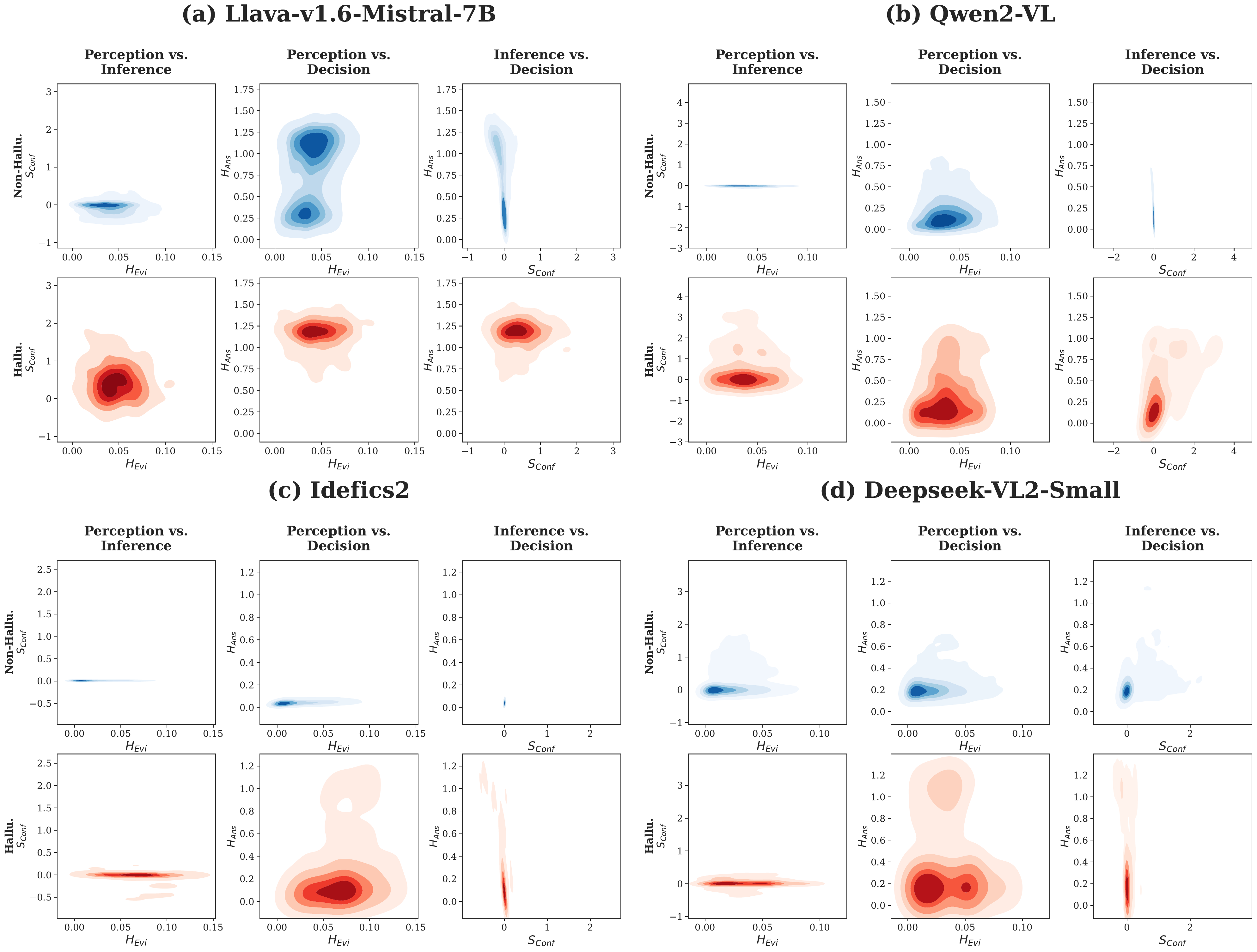}
    \caption{\textbf{Visualizing the `Cognitive Fingerprints' of Hallucination.} Density projections of the 3D Cognitive State Space, separated into non-hallucinatory (blue, top row of each pair) and hallucinatory (red, bottom row of each pair) processes. These manifolds reveal unique failure signatures for each model.}
    \label{fig:density_manifolds}
    \vspace{-4mm}
\end{figure*}

The success of our CAD framework over the supervised `Supervised Probe` is a crucial finding, suggesting that analyzing the entire cognitive trajectory for \textit{anomalous patterns} provides a richer signal than a localized biopsy of a single token's state. While the SOTA results are compelling, a deeper question arises: \textit{what underlying mechanistic differences cause performance to vary so dramatically across models?} Our framework's primary strength is its diagnostic capability. By projecting the generative process into the 3D Cognitive State Space, we can visualize the `cognitive manifolds' of nominal and hallucinatory behavior. As \cref{fig:density_manifolds} reveals, different VLM architectures follow strikingly different pathological pathways.

\paragraph{Idefics2's `Structural Disorder' Pattern.} Idefics2 presents a stark signature. Its nominal processes form an extremely compact, low-variance manifold—a tight blue cluster representing a rigid, stable cognitive workflow. Hallucinations are characterized as anomalous deviations from this stable state. This suggests a \textbf{structural failure} mechanism, explaining why our density-based anomaly detector is exceptionally effective for this model.

\paragraph{Llava's `Transparent Struggle' Pattern.} Llava-v1.6 exhibits a cognitively transparent pattern. Its hallucinatory manifold (red) occupies a region largely separable from the nominal one (blue), characterized by high Inferential Conflict and Decisional Ambiguity. Its internal cognitive struggle is explicitly manifested through our metrics, making its failure mode highly transparent.

\paragraph{Qwen2-VL \& DeepSeek's `Entangled States' Pattern.} These models exhibit the most insidious pattern: `confident lies.' The `Entangled States' pattern provides a direct, geometric explanation for their lower AUC scores. For these models, the hallucinatory manifold (red) forms its own dense, confident clusters that deeply intertwine with the 'healthy' manifold (blue). This is not a failure of our method, but a profound diagnostic finding. It reveals that for certain architectures, hallucination is not merely a process anomaly but can be a content error originating from a seemingly normal process, highlighting a key challenge for future research.

\subsection{Ablation Study and Real-World Robustness}
\label{sec:ablation}

To validate our framework's multi-component design and its practical viability, we conducted comprehensive ablation studies and stress-testing.

\begin{wrapfigure}{r}{0.55\columnwidth}
    \centering
    \includegraphics[width=\linewidth]{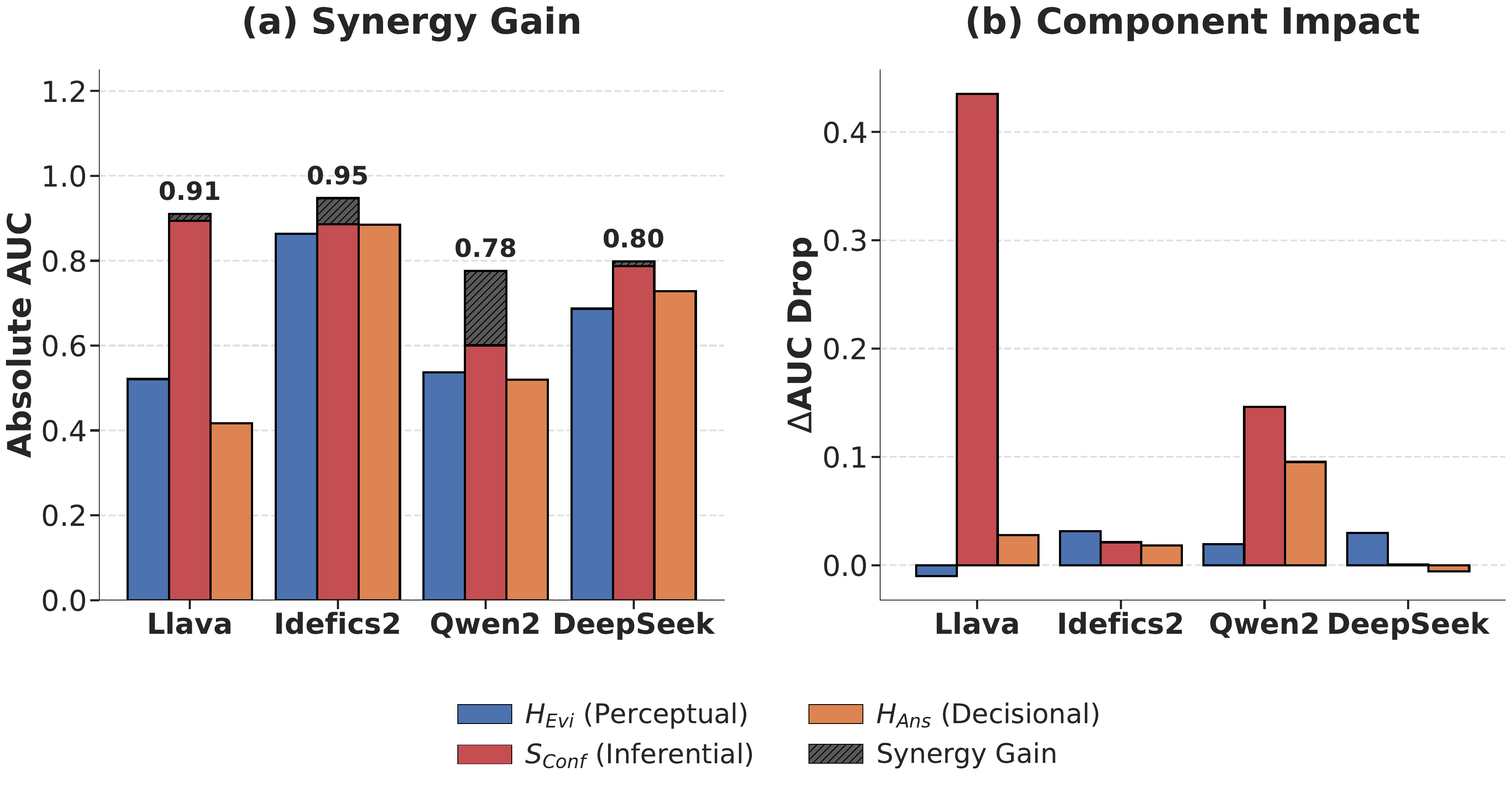}
    \caption{\textbf{Ablation Study.} (a) Standalone metrics vs. synergistic gain. The gray hatched area represents the \textit{Synergy Gain}. (b) Impact of component removal ($\Delta$AUC), revealing model-specific `diagnostic fingerprints'.}
    \label{fig:ablation_all}
    \vspace{-4mm} 
\end{wrapfigure}
\paragraph{The Necessity of a Holistic Diagnosis.} The significant `Synergy Gain' revealed in \cref{fig:ablation_all}(a) is a testament to the multi-dimensional nature of cognitive failure. For a model like Idefics2, no single probe provides a sufficient signal (individual AUCs $\sim$0.75-0.78). Only by viewing the cognitive trajectory as a point in our 3D state space can its `anomalous deviations' be reliably detected, as evidenced by the jump to 0.947 AUC, quantitatively proving that a holistic diagnosis is necessary.

\paragraph{The Adaptive Diagnostician.} \Cref{fig:ablation_all}(b) reveals the unique `diagnostic fingerprint' of each model. Our GMM-based detector acts as an adaptive diagnostician: for Llava, it learns to be highly sensitive to deviations along the \textit{Inferential Conflict} axis; for DeepSeek-VL, it identifies \textit{Perceptual Instability} as the key symptom. The GMM's ability to learn where to focus—and what to ignore (e.g., the confounding noise from $H_{\text{Evi}}$ for Idefics2)—demonstrates its power to adapt to each model's unique cognitive fingerprint.

\begin{figure}[t!]
    \centering
    \includegraphics[width=\columnwidth]{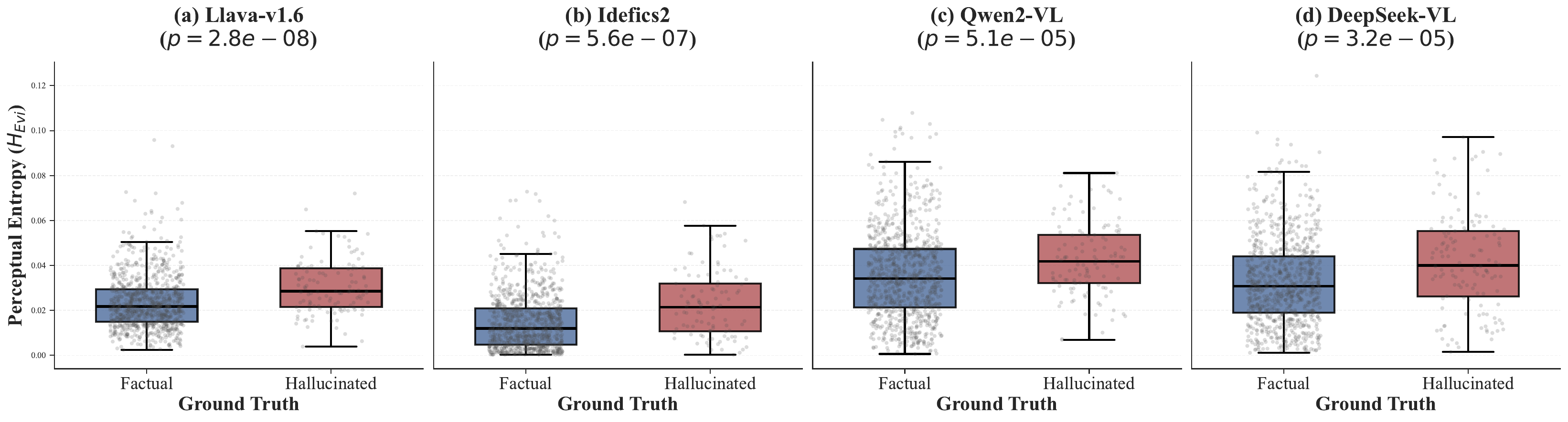}
    \caption{\textbf{Generalization of Perceptual Instability ($H_{\text{Evi}}$) to Open-Ended Captioning (MS-COCO, $N=1000$).} 
    Without any task-specific tuning, our perceptual probe consistently assigns significantly higher entropy to hallucinatory captions (red) compared to factual ones (blue) across all four architectures.
    \textbf{Statistical Significance:} The distinction is profound, with Welch's t-test yielding $p \ll 0.001$ in all cases, validating that $H_{\text{Evi}}$ captures a fundamental cognitive signature of hallucination beyond VQA formats.
    \textbf{Cross-Model Insight:} The shared Y-axis highlights that models like Qwen2-VL and DeepSeek-VL2 (bottom row) exhibit higher baseline entropy in their factual generations compared to Llava and Idefics2.}
    \label{fig:coco_boxplot}
    \vspace{-4mm}
\end{figure}

\paragraph{Generalization of Individual Probes: $H_{\text{Evi}}$ on Open-Ended Tasks.}
While our holistic framework is powerful, it is crucial to validate the independent efficacy and generalizability of its constituent probes. We conducted a targeted ablation on our \textbf{Perceptual Instability ($H_{\text{Evi}}$)} probe by evaluating it on the challenging open-ended task of MS-COCO image captioning ($N=1000$). Since unconstrained captioning lacks a binary decision boundary (e.g., Yes/No) to anchor the Inferential Conflict ($S_{\text{Conf}}$) and Decision Entropy ($H_{\text{Ans}}$) metrics, this setting allows us to isolate perceptual drift as a primary driver of hallucination in free-form generation. As shown in \cref{fig:coco_boxplot}, $H_{\text{Evi}}$ scores are significantly higher for hallucinated captions (validated via CHAIR~\cite{rohrbach2018object}) across all models, with profound statistical significance ($p \ll 0.001$). This result not only demonstrates the standalone power of our perceptual probe but also proves its ability to generalize far beyond structured VQA tasks. 

\begin{figure}[t!]
    \centering
    \includegraphics[width=\columnwidth]{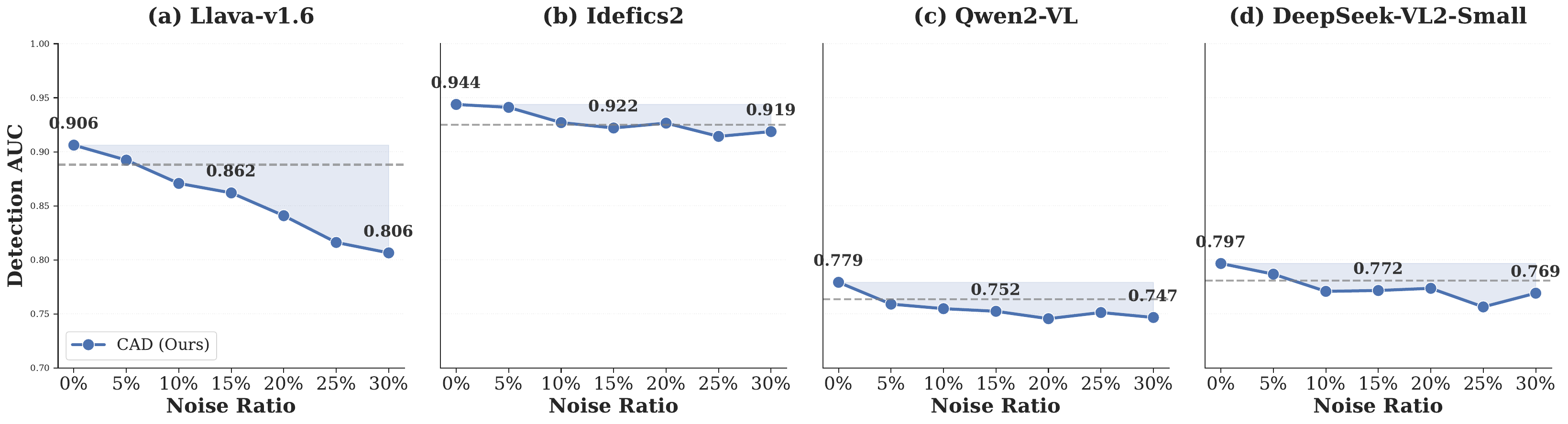}
    \caption{\textbf{Robustness to Calibration Contamination across Architectures.} 
    We evaluate detection performance (AUC) as the calibration set is increasingly contaminated with hallucinatory samples (0\%--30\%). 
    \textbf{(a) Sensitivity of Transparent Struggle:} Llava-v1.6 exhibits a noticeable drop, as mistakenly treating its highly conflicted, high-variance hallucinatory states as nominal forces the GMM to blur the anomaly boundary.
    \textbf{(b) Resilience of Structural Disorder:} Idefics2 maintains high performance ($>0.91$) with negligible degradation, as its compact nominal core remains robust against geometrically distinct outliers.
    \textbf{(c, d) Stability of Entangled States:} Qwen2-VL and DeepSeek-VL2 show flatter trajectories; their performance is constrained by the intrinsic geometric overlap of their states rather than calibration noise.
    The gray dashed line shows a strict 2\% performance drop threshold relative to the clean baseline.}
    \label{fig:robustness_noise}
    \vspace{-4mm}
\end{figure}

\paragraph{Robustness to Calibration Contamination.}
CAD's reliance on weak supervision—calibrating solely on samples with correct final answers—is a significant practical advantage, as perfectly clean data is often unavailable in real-world deployments. To evaluate its limits, we stress-test the framework by intentionally contaminating the calibration set with 0\%--30\% undetected hallucinations. As illustrated in \cref{fig:robustness_noise}, the resilience of our GMM-based detector aligns elegantly with the underlying `cognitive fingerprints' of each architecture. Idefics2 remains remarkably robust, maintaining an AUC above 0.91 even under 30\% contamination. This confirms its \textit{Structural Disorder} profile: the nominal cognitive core is so geometrically compact that the GMM effectively treats injected hallucinations as negligible outliers, leaving the learned density of the manifold intact. 

In contrast, Llava-v1.6's noise-sensitivity reflects its \textit{Transparent Struggle}: including high-variance hallucinations forces GMM variance dilation, blurring thresholds. As its hallucinations are characterized by conspicuously high-variance and extreme inferential conflict, mistakenly including them as ``nominal'' forces the GMM to artificially stretch its variance parameters to encompass these anomalies. Conversely, Qwen2-VL and DeepSeek-VL2 remain stable as noise merely reinforces their inherent \textit{Entangled States} overlap. Such resilience under severe contamination underscores CAD's utility for robust real-world auditing.

\section{Conclusion}
\label{sec:conclusion}
In this paper, we introduce a diagnostic framework that reframes VLM hallucination from a static output error to a dynamic failure within the model's cognitive process. By modeling generation as a three-stage cognitive trajectory and projecting it into an interpretable state space, we develop a state-of-the-art, single-pass, and weakly supervised anomaly detector. Crucially, our robust framework provides the first mechanistic categorization of VLM hallucinations, uncovering distinct `cognitive fingerprints' for different architectures—from transparent failures to deeply entangled errors. This approach advances beyond mere detection, enabling the stage-by-stage diagnosis of complex, multi-modal failure modes, ultimately providing a transparent new lens for auditing and building reliable Vision-Language Models.

{
    \small
    \nocite{*}
    \bibliographystyle{template}
    \bibliography{main}
}

\appendix

\section{Appendix}
\label{sec:appendix}

\subsection{Implementation and Reproducibility Details}
\label{app:implementation}

\paragraph{Prompt Template.}
For all experiments, we used a consistent Chain-of-Thought prompt to elicit the three-stage cognitive process. The template enforces a structured output format to facilitate parsing of the evidence chain ($\mathcal{T}_{\text{evi}}$) and the final answer ($\mathcal{A}$).

\begin{figure}[h]
\centering
\begin{lstlisting}[basicstyle=\small\ttfamily, frame=single, breaklines=true]
User: <image>
You are an expert image analyst. Follow this output format strictly:
First, provide a brief explanation of what 
you see in the image. Then, conclude with 
'Therefore, the final answer is Yes.' or 
'Therefore, the final answer is No.'
Question: {question}

Assistant:
\end{lstlisting}
\caption{The Chain-of-Thought prompt template used in our experiments. The explicit structural constraints ensure reliable separation of evidence and decision tokens.}
\label{fig:prompt_template}
\end{figure}

\paragraph{Metric Calculation Details.}
\label{app:metric_calc}
The three core metrics are calculated as follows. Let $\mathcal{V}$ be the vocabulary and $\mathcal{I}$ the image. 

\textbf{1. Uncertainty Word Set ($\mathcal{V}_U$).} The complete set of uncertainty-related words used to construct $\mathcal{V}_U$ was adapted from prior work on linguistic uncertainty~\cite{Ji2025Calibrating, yona2024can}: 
\texttt{\{"probably", "likely", "possibly", "might", "may", "seems", "appears", "perhaps", "suggests", "could", "believe", "guess", "assume", "unlikely", "not sure", "could be"\}}, including their space-prefixed variations (e.g., " probably").

\textbf{2. Probe Calculations.}
\begin{itemize}[leftmargin=*]
    \item \textbf{$H_{\text{Evi}}$ (Perceptual Instability)}: For each token $i$ in the evidence chain $\mathcal{T}_{\text{evi}}$ with probability distribution $p_i$, let the total probability of uncertainty tokens be $P_i(U) = \sum_{t \in \mathcal{V}_U} p_i(t)$. The semantic binary entropy is $H_{B}(p_i) = -P_i(U)\log_2 P_i(U) - (1-P_i(U))\log_2 (1-P_i(U))$. $H_{\text{Evi}}$ is the mean of $H_B(p_i)$ over all tokens in $\mathcal{T}_{\text{evi}}$.
    
    \item \textbf{$S_{\text{Conf}}$ (Inferential Conflict)}: Let $t_{\text{ans}}$ be the final answer token (e.g., "Yes" or "No") and $\mathcal{C}_{\text{evi}}$ be the full context preceding it. This is calculated as the Conditional Pointwise Mutual Information (CPMI):
    \begin{equation}
        S_{\text{Conf}} = \log P(t_{\text{ans}} | \mathcal{I}, \mathcal{C}_{\text{evi}}) - \log P(t_{\text{ans}} | \emptyset_{\mathcal{I}}, \mathcal{C}_{\text{evi}})
    \end{equation}
    The second term is obtained by a \textit{teacher-forcing replay} of the generated token sequence using only the textual prompt as input, with visual features entirely excluded (passed as zero tensors or `None`). This ratio quantifies the specific information gain from the visual modality at the decision step.
    
    \item \textbf{$H_{\text{Ans}}$ (Decisional Ambiguity)}: This is the standard Shannon entropy of the probability distribution $P$ over the entire vocabulary $\mathcal{V}$ at the position of the final answer token: $H_{\text{Ans}} = - \sum_{w \in \mathcal{V}} P(w)\log P(w)$.
\end{itemize}

\paragraph{Supervised Probe Implementation.} For the `Supervised Probe` baseline, we used the code provided by~\cite{Chen2024INSIDE} to implement a linear probe with `LogisticRegression` from `scikit-learn`. The probe was trained on hidden states extracted from the final token of the generated evidence chain. This location was determined by identifying the sequence marker "Therefore, the final answer is" and taking the hidden state of the token immediately preceding it. Training was performed on a balanced dataset using 5-fold stratified cross-validation to ensure fair comparison.

\subsection{GMM Calibration and Hyperparameters}
\label{app:gmm_details}

\paragraph{Model Selection via BIC.}
A critical hyperparameter for our Gaussian Mixture Model (GMM) is the number of components $K$. We assume the "manifold of nominal cognition" is not necessarily unimodal (e.g., different reasoning types may form different clusters). To determine the optimal $K$, we employed the Bayesian Information Criterion (BIC) on the calibration set. We swept $K \in [1, 10]$.

\begin{figure}[h]
    \centering
    \includegraphics[width=0.95\columnwidth]{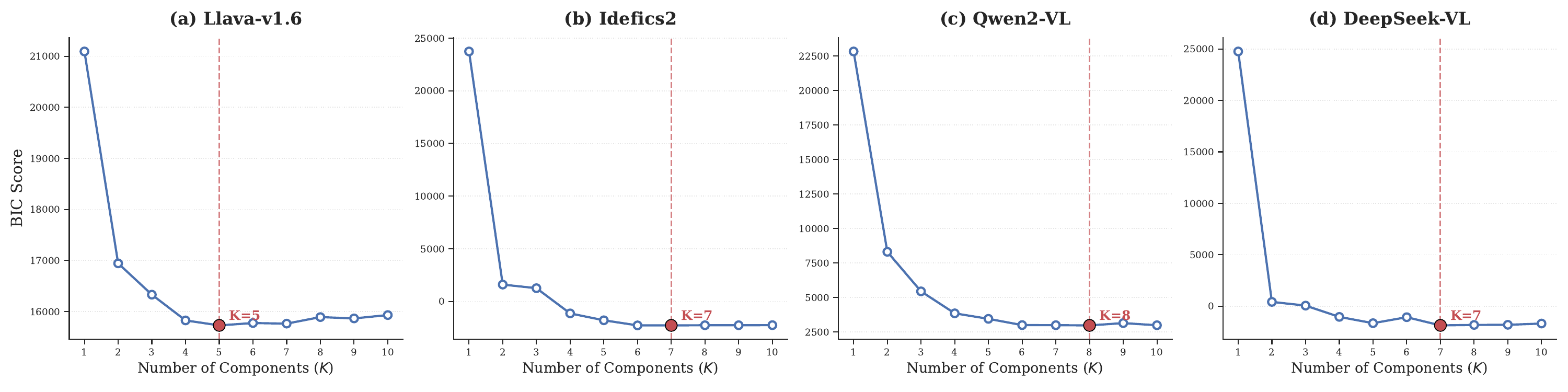}
    \caption{\textbf{GMM Model Selection.} Bayesian Information Criterion (BIC) scores for varying numbers of Gaussian components. The red dashed line indicates the selected optimal $K$ for each model, minimizing the BIC score.}
    \label{fig:bic_curves}
\end{figure}

As shown in \cref{fig:bic_curves}, distinct minima were observed for each model, reflecting their varying internal latent structures. The optimal components selected were: \textbf{5 for Llava-v1.6}, \textbf{7 for Idefics2}, \textbf{8 for Qwen2-VL}, and \textbf{7 for DeepSeek-VL2}.

\paragraph{Calibration Data Purification (Coherence Filter).}
To ensure our GMM models the density of \textit{truly} nominal cognitive processes, we applied a "Coherence Filter" to the calibration set (which consists of samples where the final answer is correct). This step removes "lucky guesses"—cases where the model answers correctly despite hallucinating in the evidence chain.
The heuristic rules are:
\begin{enumerate}
    \item \textbf{Negation Consistency:} If the final answer is "Yes", the evidence chain must not contain strong negation phrases (e.g., "no [object]", "not present") associated with the query object.
    \item \textbf{Object Existence:} If the final answer is "Yes", the object name must appear in the evidence chain without being preceded by negative modifiers.
\end{enumerate}
Samples failing these checks were excluded from the calibration set $\mathcal{D}_{\text{cal}}$, ensuring the learned manifold $\mathcal{S}_{\text{nominal}}$ represents consistent reasoning trajectories.

\section{Future Work}
\label{sec:limitations}
The diagnostic nature of our framework opens several exciting avenues for future work. The ability to pinpoint the stage of cognitive failure paves the way for \textbf{targeted, stage-specific mitigation strategies}. For instance, a high $H_{\text{Evi}}$ could trigger a visual re-evaluation mechanism (e.g., zooming or re-sampling), while a high $S_{\text{Conf}}$ could activate a more rigorous logical verification step (e.g., self-consistency checks). Furthermore, extending this framework to mitigate the most deceptive `deep cognitive errors'—where models are confidently wrong (low entropy, high distance)—remains a critical direction for building trustworthy VLM systems.
\clearpage

\end{document}